\documentclass{article}
\usepackage{amssymb}

\usepackage[preprint]{corl_2025} 
\usepackage{graphicx}
\usepackage{wrapfig}
\usepackage{amsmath}

\usepackage{geometry}
\usepackage{longtable}
\usepackage{booktabs}
\usepackage{adjustbox}
\usepackage{multirow}
\usepackage{xspace}          
\usepackage{siunitx}         
\usepackage{enumitem}    

\setlist[itemize]{leftmargin=1.1em}
\setlist[enumerate]{leftmargin=1.5em}

\newcommand{\TODO}[1]{\textbf{\color{red}[TODO: #1]}}

\newcommand{\methodname}{\texttt{SSR}}
\newcommand{\methodnamefl}{\texttt{SSR-FL}}
\newcommand{\methodfullname}{Similarity Space Replication}

\renewcommand{\TODO}[1]{}

\title{\methodname: A Generic Framework for Text-Aided Map Compression for Localization}

\author{
  Mohammad Omama$^{1}$, 
  Po-han Li$^{1}$, 
  Harsh Goel$^{1}$, 
  Minkyu Choi$^{1}$, 
  Behdad Chalaki$^{2}$, \\
  \textbf{ 
  Vaishnav Tadiparthi$^{2}$, 
  Hossein Nourkhiz Mahjoub$^{2}$, Ehsan Moradi Pari$^{2}$, Sandeep P. Chinchali$^{1}$} \\
  $^{1}$The University of Texas at Austin, $^{2}$Honda Research Institute \\
}

\begin{document}
\maketitle


\begin{abstract}

Mapping is crucial in robotics for localization and downstream decision-making. As robots are deployed in ever‑broader settings, the maps they rely on continue to increase in size. However, storing these maps indefinitely (cold storage), transferring them across networks, or sending localization queries to cloud-hosted maps imposes prohibitive memory and bandwidth costs. 
We propose a text-enhanced compression framework that reduces both memory and bandwidth footprints while retaining high-fidelity localization. The key idea is to treat text as an alternative modality—one that can be losslessly compressed with large language models. We propose leveraging lightweight text descriptions combined with very small image feature vectors, which capture \emph{``complementary information"} as a compact representation for the mapping task. 
Building on this, our novel technique, \methodfullname\ (\methodname), learns an adaptive image embedding in one shot that captures only the information \emph{``complementary"} to the text descriptions.
We validate our compression framework on multiple downstream localization tasks, including Visual Place Recognition as well as object-centric Monte-Carlo localization in indoor as well as outdoor settings. \methodname\ achieves $2 \times$ better compression than competing baselines on state-of-the-art  datasets, including TokyoVal, Pittsburgh30k, Replica, and KITTI.
\end{abstract}
\vspace{-1em}



\begin{figure*}[h]
    \centering
    \includegraphics[width=\textwidth]{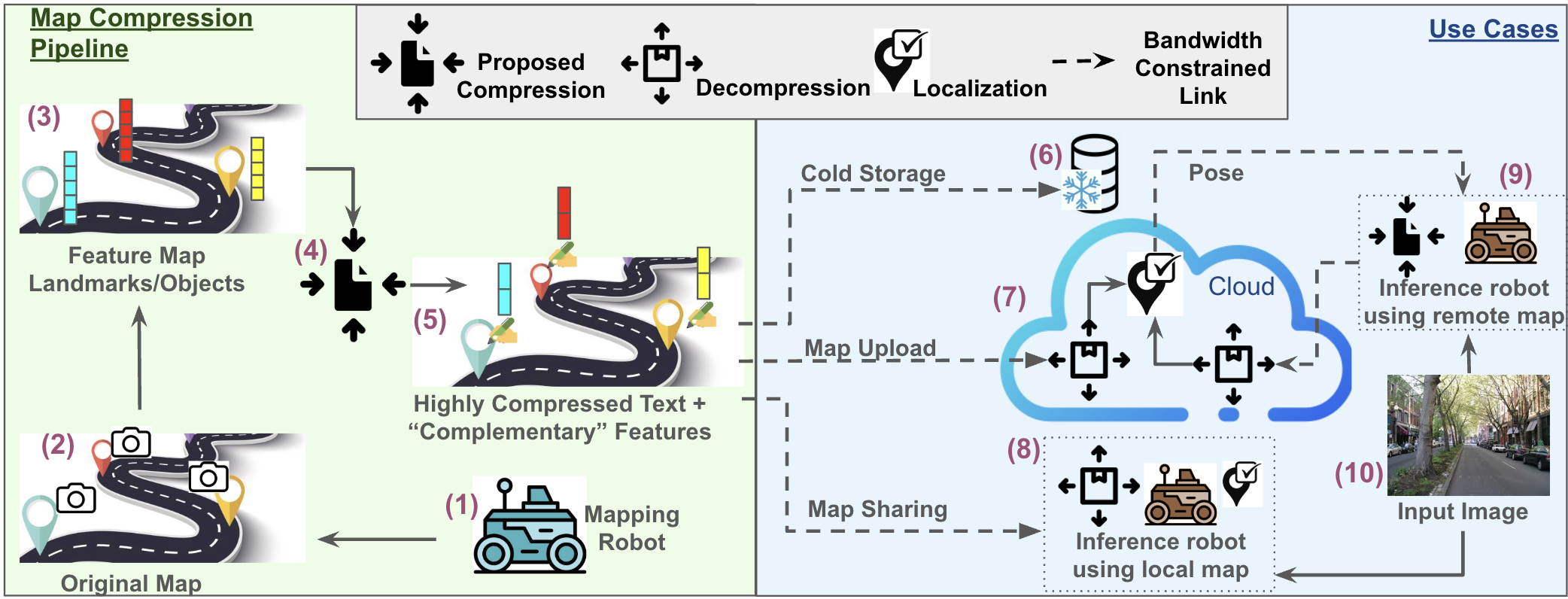} 
    \caption{
    \small\textbf{Map compression pipeline and downstream use-cases.} The mapping robot (1) first creates a standard map of the environment (2), which is then converted into a feature map (3). Our proposed compression framework (4) processes these feature maps into highly compressed text and complementary feature vectors (5). The compressed map can be stored in cold storage (6), uploaded to a server (7) for remote queries, or transmitted to another robot (8). During inference, if the robot has the compressed map locally (8), standard localization is performed. Otherwise, the input image (10) is compressed by the robot (9), sent to the server, decompressed, and localized, with the resulting pose returned to the robot.
    }

    \label{fig:system} 
\end{figure*}

\section{Introduction} \label{sec:intro}

\begin{wrapfigure}{r}{0.47\textwidth}
    \vspace{-1em}
  \centering
  \includegraphics[width=0.47\textwidth]{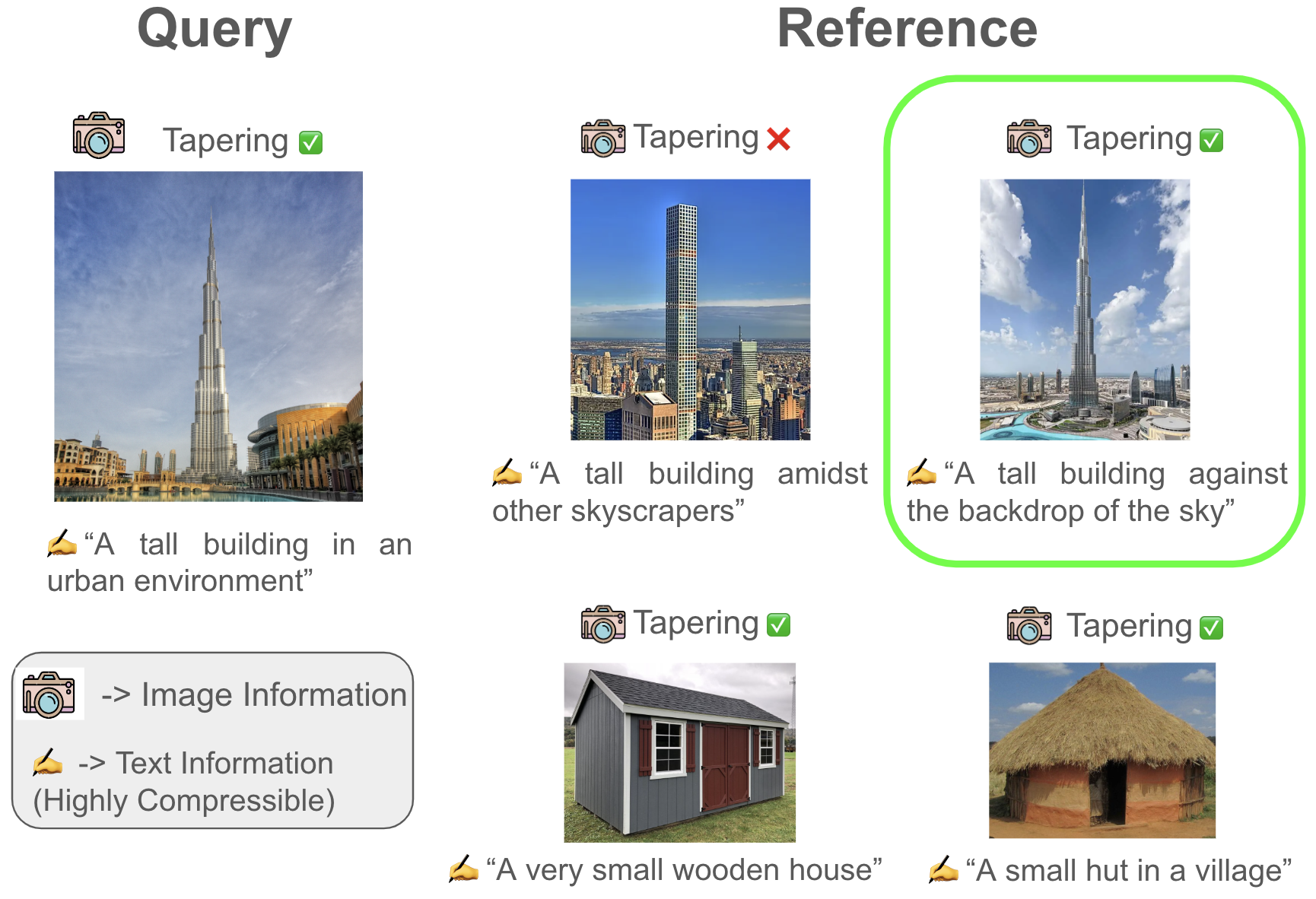}
  \caption{
  \small\textbf{Highly compressible text descriptions combined with ``complementary" image information are enough for effective localization.} Text descriptions (highly compressible) are good enough to discard the bottom two candidates in the reference set for the given query but struggle to distinguish between the top two. Integrating complementary details from the image, like whether the building tapers, ensures a precise match.}
  \label{fig:teaser}
\end{wrapfigure}
Modern robots—like self-driving cars, delivery drones, and warehouse vehicles—localize by matching live sensory input against \emph{ever-growing maps}. These maps grow continuously as robots are deployed to new areas. In city-scale ride-hailing systems, for example, the landmark database can exceed terabytes of data. Sending daily map updates to hundreds of cars can overwhelm cellular networks.
Bandwidth and storage constraints appear in other settings as well. An indoor service robot frequently sends image snippets to a cloud server that hosts a high-resolution map of its operational area; each query incurs both latency and bandwidth costs. Further, survey drones that fly week-long missions over farmland collect petabytes of imagery; archiving this data for later use has prohibitive memory costs. All of these scenarios underline a simple point: without effective compression techniques, map storage and communication quickly become the dominant costs in robotic localization.
Consequently, there is a pressing need for map compression techniques tailored for localization that \textbf{a) reduce memory and bandwidth footprints} and \textbf{b) seamlessly adapt to varying resource constraints.}

To address this, we propose a novel map compression pipeline (illustrated in Fig.~\ref{fig:system}), which transforms feature maps into highly compressed text descriptions and complementary feature vectors, enabling efficient storage, transmission, and localization across diverse deployment settings.
Existing classical and neural compression techniques focus on reconstruction quality, making them unsuitable for localization. Other popular map compression techniques primarily focus on dimensionality reduction or quantization.

Contrastingly, our approach, leverages lightweight, easily-compressible text descriptions generated by vision-language models, combined with compact image feature vectors. These compact feature vectors are learned using our novel method, \methodfullname\ 
(\methodname), and capture ``complementary information". 
Together, the text descriptions and complementary image vectors enable a compact solution for map compression tailored for localization. 
Fig. \ref{fig:teaser} shows a Visual Place Recognition (VPR) scenario where the goal is to find the closest match for a query image of a building. Text descriptions can easily eliminate the bottom two candidates in the reference set, but fail to distinguish between the top two. By integrating text with ``complementary features"—like whether the building tapers—we accurately identify the correct match using minimal data.
Moreover, our approach \methodname\ is adaptive and does not require separate training to adapt to fluctuating bandwidth.


\textbf{Motivation for Using Text:} Our two key insights are that text is inherently more compact and significantly easier to compress than images or feature vectors. For example, a JPEG image~\cite{wallace1991jpeg} is typically around 500~KB, while a CLIP feature vector~\cite{radford2021learning} is approximately 4~KB. In contrast, a concise 1--2 line caption describing the image is roughly 0.1~KB. Moreover, leveraging recent advancements such as LLMZip~\cite{valmeekam2023llmzip}, the caption can be further compressed losslessly---down to approximately 0.025~KB---making text an ideal modality for compression. As a result, we only need marginal or complementary information from the image, which we obtain using our novel technique, \methodname. To the best of our knowledge, this is the first work to demonstrate that recent LLM-based compression techniques can be extended to robotic map compression.

\begin{figure*}[t]
    \centering
    \includegraphics[width=\textwidth]{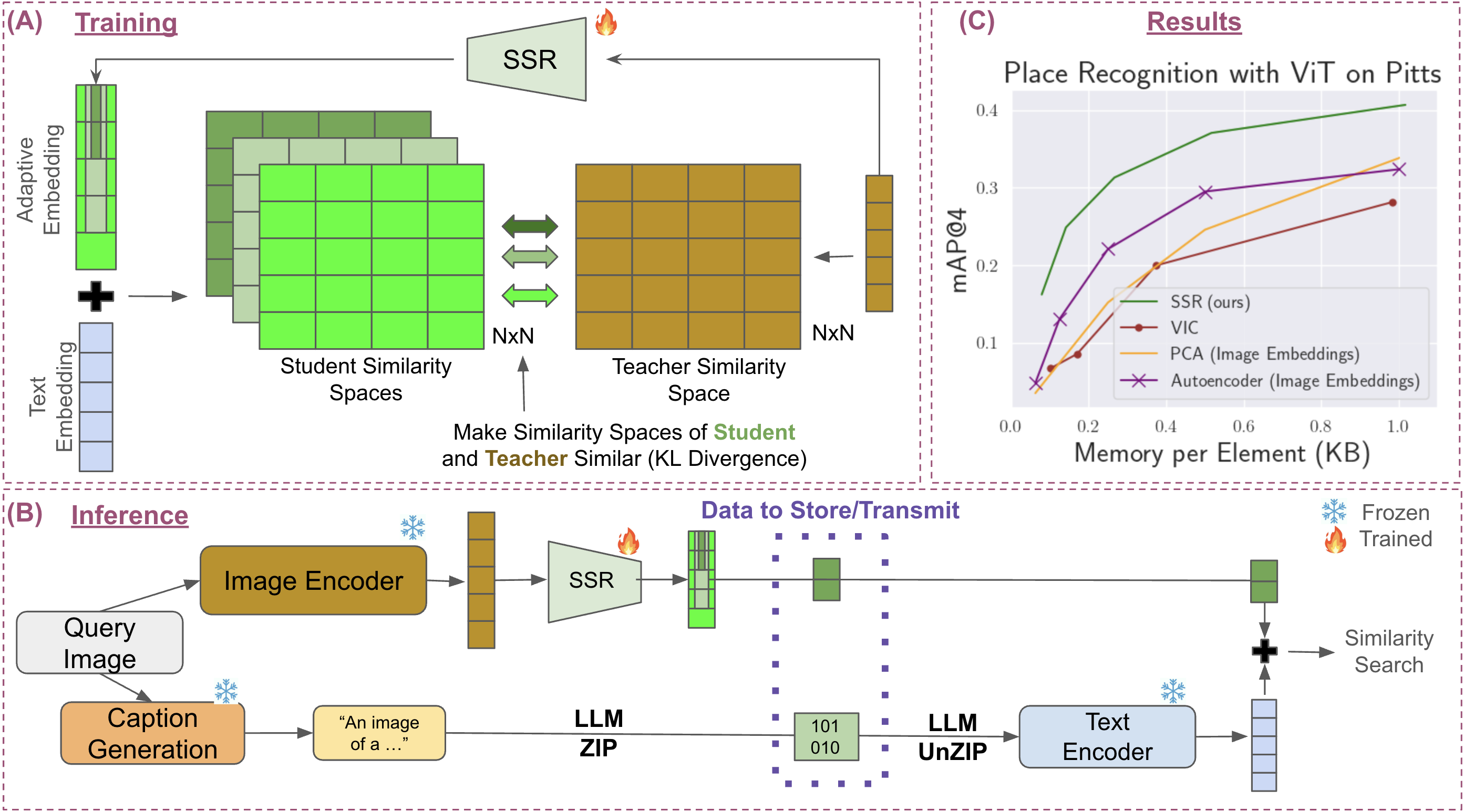} 
    \caption{
    \small\textbf{Pipeline for \methodname.} \textbf{(A)} \methodname\ (detailed in Sec. \ref{sec:method_ssr}) learns adaptive embeddings (green) from the image embeddings (brown) that capture complementary information to the text embeddings (blue). \textbf{(B)} During inference, \methodname\ projects the image embedding to any desired dimension based on the constraints. In parallel, captions are generated (Sec. \ref{sec:method_vlm}) and then compressed with LLMZip (Sec. \ref{sec:method_llmzip}). The projected complementary feature vector and LLMZipped text are then stored or transmitted. \textbf{(C)} \methodname\ (green) outperforms all competing compression baselines (complete results in Sec. \ref{sec:exp_results}).
    }
    \label{fig:pipeline} 
\end{figure*}

The central idea of this work is presented in Fig. \ref{fig:pipeline}. To summarize, our contributions are:
\begin{enumerate}
    \item We propose a novel technique that uses text, highly compressed with LLMZip, and complementary image information to achieve effective map compression.
    \item Our method, \methodfullname\ (\methodname), effectively learns an adaptive embedding that captures information complementary to text, as shown in Fig. \ref{fig:pipeline}(A) and further discussed in Sec. \ref{sec:method_ssr}. \methodname\ can work with any feature extractor as demonstrated in Sec. \ref{sec:exp_results}.
    \item Our approach achieves $2 \times$ better compression than competing baselines on state-of-the-art (SOTA) localization datasets, as shown in Fig. \ref{fig:pipeline}(C) and discussed in Sec. \ref{sec:exp_results}. We also release our codebase in the supplementary material.
\end{enumerate}

Throughout the paper, when we refer to \methodname\ in the context of methodology, we mean our approach for learning complementary image feature factors, as illustrated in Fig. \ref{fig:pipeline}(A). In the context of experiments, these terms denote the complete compression setup, incorporating both the complementary image feature vectors and the LLMZipped text during inference as shown in Fig. \ref{fig:pipeline}(B).

\section{Background and Related Work}

\textbf{Deep Metric Learning:}
Deep Metric Learning (DML) provides a foundation for image retrieval \cite{dml-survey}, focusing on generating latent representations where similar items are positioned closely in the embedding space, while dissimilar items are pushed apart. Numerous techniques have been developed \cite{dml1, dml2facenet, dml3, dml4-npair, dml5-msl, omama2024exploiting, li2024csa} to create these representations, leveraging strategies like contrastive loss \cite{dml-contrastive}, triplet loss \cite{dml2facenet}, advanced N-pair loss \citep{dml4-npair}, and multi-similarity loss \citep{dml5-msl}. Similarly, VPR applies DML principles to embed images in a way that clusters similar locations together while distancing unrelated ones \cite{vpr-intro}. 
For VPR, specific training objectives have been designed \cite{vpr1, vpr2, vpr3, cosplace} to enhance retrieval capabilities by refining these embeddings to improve the clustering of like locations and the separation of distinct ones. 
Furthermore, composed retrieval \cite{composedret1, composedret2} involves using a combination of text, images, and sketches to retrieve images.
However, while these approaches focus on learning effective latent representations for retrieval tasks, they do not address the challenge of compression, particularly in a localization setting.

\textbf{Image Compression Techniques}: Traditional image compression methods, such as JPEG \cite{wallace1991jpeg} and JPEG2000 \cite{skodras2001jpeg2000}, rely on mathematical transformations and quantization processes to reduce file sizes while preserving visual quality. 
Advancements in Deep Learning (DL) have led to more effective image compression methods \cite{mishra2022deep, ma2019image}. Autoencoders (AEs) are popular for this task \cite{theis2017lossy}, and Variational Autoencoders (VAEs) extend AEs with a probabilistic framework for more flexible latent representations \cite{kingma2013auto}. Generative adversarial networks (GANs) further enhance perceptual quality in image reconstruction \cite{agustsson2019generative}. Recent methods also use arithmetic encoding for more efficient latent space compression \cite{cheng2020, vicvariationalimagecompressionscale}. Additionally, quantization techniques reduce file sizes by mapping pixel values or latent representations to finite sets \cite{toderici2016full}. However, these methods are generally reconstruction-focused, rather than optimized for retrieval.
For retrieval-oriented compression, methods typically employ autoencoders \cite{kingma2013auto} or PCA \cite{abdi2010principal} for dimensionality reduction of the feature vector used in retrieval, or quantization \cite{gersho2012vector} to reduce its size. These methods, however, tend to degrade performance significantly at higher compression levels. In contrast, \methodname\ introduces a novel approach that leverages text captions, combined with complementary image embeddings, to enable memory-optimized retrieval.

\textbf{Vision Language Models (VLMs):}
VLMs such as InternVL \cite{chen2024internvl}, LLaVA \cite{liu2023llava}, BLIP-2 \cite{li2023blip2} and InstructBlip \cite{dai2023instructblipgeneralpurposevisionlanguagemodels} have gained popularity in recent years for tasks such as VQA, multi-modal dialogue generation, image captioning, etc. We leverage these recent advancements and apply them in the image retrieval domain by relying on the zero-shot image captioning capabilities of VLMs.

\textbf{Similarity Preserving Knowledge Distillation:}
Much like traditional distillation techniques \citep{dist1, dist2, dist3, dist4, dist5, dist6}, similarity-preserving knowledge distillation (SPKD) focuses on transferring knowledge from a teacher model to a student model while ensuring that the similarity relationships between data points in the feature space are retained. 
FitNets \citep{romero2014fitnets} guide the student model using the teacher's intermediate features, while attention transfer aligns attention maps \citep{dist4}. Pairwise similarity preservation \citep{tung2019similarity}, Relational Knowledge Distillation (RKD) \citep{park2019relational}, and Contrastive Representation Distillation (CRD) \citep{tian2019contrastive} retain data relationships, with CRD using contrastive loss to maximize mutual information. We adapt SPKD to generate ``complementary" embeddings from image feature extractors. These embeddings, when combined with text embeddings, preserve the original feature extractor's similarity relationships.

\textbf{Adaptive Feature Embeddings:}
A fundamental challenge in representation learning is the need for adaptive embedding sizes to accommodate different compute/memory limits. Prior works have developed adaptive feature embeddings for tasks such as reconstruction \citep{li2024task} and classification \citep{kusupati2022matryoshka}. In contrast, our approach learns adaptive embeddings for localization-focused map compression.
\section{Methodology} \label{sec:method}



\subsection{Caption Generation Using Vision-Language Models} \label{sec:method_vlm}
Our approach requires a modality switch, necessitating a caption generation module, for which any  suitable  VLM could be used. We use a VLM called LLaVA (Large Language and Vision Assistant) \cite{liu2023llava}, a multimodal framework that integrates Large Language Models (LLMs) with visual encoders to enable image-captioning capabilities. LLaVA generates detailed, context-rich captions using a Vision Transformer (ViT) to extract image features, which are then interpreted by an LLM to create descriptions.
In our experiments, we employ the LLaVA-13B model and consistently use the prompt “describe the image in two lines” to guide caption generation. While we can use any caption generation model, using LLaVA provides unique benefits. As a foundation model, LLaVA is versatile and can be applied across diverse datasets without retraining. Additionally, being based on an LLM, LLaVA brings further advantages like LLM-based compression as discussed in Sec. \ref{sec:method_llmzip}.

\subsection{Extreme Lossless Text Compression} \label{sec:method_llmzip}
With the rise of LLMs, new lossless extreme text compression techniques have emerged, making text an ideal modality for efficient compression. In this context, we leverage LLMZip \cite{valmeekam2023llmzip}, a SOTA compression method that harnesses the power of LLMs to perform lossless text compression. By utilizing the predictive capabilities of these models, LLMZip encodes text more efficiently, transforming token predictions into highly compact bit representations.
LLMZip requires some LLM to work with. It was originally designed for LLaMA-7B \cite{touvron2023llama}. Since we are already using LLaVA for captioning and LLaVA is based on an LLM, we directly integrate it with the LLMZip framework, eliminating the need for an additional LLM for probability predictions. Following LLMZip's pipeline, LLaVA’s generated captions undergo tokenization and probability estimation, with predicted tokens ranked by likelihood and compressed using arithmetic coding. By using LLMZip in the loop, a lot of information can be stored or transmitted as highly compressed text while only some additional information is needed from the image vectors. 

\subsection{Learning Complementary Information Using \methodfullname\ (\methodname)} \label{sec:method_ssr}

\noindent \textbf{Overview:} Fig. \ref{fig:pipeline}(A) illustrates the central idea behind \methodname. Let ${z} \in \mathbb{R}^d$ represent the full image embedding vector generated from any feature extractor of choice. Our goal is to reduce the dimensionality of ${z}$ such that the performance of this reduced embedding, when combined with the text embedding, approximates that of the full image embedding. In essence, the complete image embedding ${z}$ guides the creation of a smaller, ``complementary" image embedding. The key insight is that the complete image embedding is well-suited for localization tasks. We create a similarity space based on this complete embedding, represented by an $N \times N$ similarity matrix, where $N$ is the number of elements in the database. We then construct a comparable similarity space using the  ``complementary" embedding in conjunction with the text embedding. Our objective is to ensure that the latter similarity space closely approximates the former. To achieve this, we introduce a Kullback-Leibler (KL) divergence loss between the two similarity matrices, similar to \cite{cvpr-context-sim-distill}, which we refer to as the \methodfullname\ loss. 
An important aspect of our approach is adaptability to varying bandwidth and memory constraints. Rather than training separate \methodname\ models for each possible embedding dimension, we train a single \methodname\ model that can produce adaptive embeddings by following the approach in \citep{kusupati2022matryoshka}, and optimizing the \methodfullname\ loss across each nested dimension. This flexibility allows us to select any number of dimensions from the embedding, with each selection providing an effective trade-off between performance and memory.

\noindent \textbf{Approach:} Let $\mathcal{C} \subset [d]$ represent a collection of embedding dimensions and define a neural network $\mathcal{G}(\cdot; \phi)$ with parameters $\phi$. The objective is to learn a mapping $\hat{z} = \mathcal{G}(z; \phi)$ such that each of the first $c$ dimensions, for $c \in \mathcal{C}$, of the resulting embedding vector $\hat{z}^{1:c} \in \mathbb{R}^c$ maintains performance for ${z}$ when combined with the text embedding ${z_{\text{text}}}$.

We first calculate a cosine similarity matrix $\mathcal{S} \in \mathbb{R}^{N \times N}$ by evaluating the cosine similarities between each pair of $z$ in the database. 
We term this matrix the cosine similarity space of the reference embeddings $Z$ or the ``teacher similarity space" as shown in Fig. \ref{fig:pipeline}(A). Let $\mathcal{S}_j$ represent the $j^{\text{th}}$ row of $\mathcal{S}$. For each $c \in \mathcal{C}$, we define a ``student similarity space" $\tilde{\mathcal{S}}^c \in \mathbb{R}^{N \times N}$ by combining $\hat{z}^{1:c}$ with the text feature vector $z_{\text{text}}$. 
We then use the KL divergence loss $l^c$ between $\tilde{\mathcal{S}}^c$ and $\mathcal{S}$ for each $c$ as:
\begin{equation}
l^c = \sum_{j=1}^N \mathrm{D}_{\mathrm{KL}}(\tilde{\mathcal{S}}^c_j \| \mathcal{S}_j).
\end{equation}
The total loss $L$ is expressed as:
\begin{equation}\small
L_{\methodname} = \sum_{c\in \mathcal{C}} l^c = \sum_{c \in \mathcal{C} } \sum_{j=1}^{N} \mathrm{D}_{\mathrm{KL}}(\tilde{\mathcal{S}}^c_j \| \mathcal{S}_j).
\end{equation}

\subsection{Applying \methodname\ in Different Localization Settings}\label{sec:adapting_loc}
\methodname\ can be seamlessly applied to VPR settings by captioning each image to obtain the corresponding text description. In this case, the number of database elements $N$ equals the number of images in the VPR database. For object-centric localization, we introduce a minor adaptation to the \methodname\ pipeline. Recent object-centric Monte-Carlo localization approaches, like ConceptGraphs~\cite{conceptgraphs} and Sparseloc~\cite{sparseloc}, extract object embeddings from multiple viewpoints and fuse them during the mapping process. Following~\cite{conceptgraphs}, we generate object captions by first cropping the object from each viewpoint and passing the crops to a captioning model. We then prompt the VLM to synthesize a unified caption across all viewpoints. In this setting, $N$ corresponds to the number of distinct objects in the map. The remainder of the map compression pipeline remains unchanged. 

\textbf{Multi-robot Settings:} 
In Appendix~\ref{app:method_ssrfl}, we demonstrate that our compression pipeline can be seamlessly extended to multi-robot federated learning settings. 
Our approach outperforms other methods in a federated setup,  
due to its better data efficiency, as further discussed in Appendix~\ref{app:dataeff}.

\section{Experiments}

\subsection{Experimental Setup}\label{sec:exp_setup}
\textbf{Datasets}: 
We validate our compression framework across multiple downstream localization tasks, including visual place recognition and object-centric Monte Carlo localization, in both indoor and outdoor environments. 
For visual place recognition, we report results on the Pittsburgh30K~\citep{pittsdataset} and TokyoVal~\citep{tokyodataset} datasets. 
For object-centric Monte Carlo localization, we present results on the Replica~\citep{replica} (indoor) and KITTI~\citep{kitti} (outdoor) datasets.


\noindent \textbf{Feature Extractors Evaluated}: To demonstrate that our approach is applicable with any choice of feature vector used for localization, we conducted experiments using three different feature extractors, each pre-trained with different datasets and techniques: DINO \citep{dino}, DINOv2 \citep{dinov2}, and ViT \citep{vit}. The ViT and DINO models were both pre-trained in a self-supervised manner on the ImageNet dataset \citep{russakovsky2015imagenet}, while the DINOv2 model was trained on the LVD-142M dataset \citep{dinov2data:lvd142m}. These feature vectors are used as teachers to train the \methodname\ model as discussed in Sec. \ref{sec:method_ssr}. 

\noindent \textbf{Settings}: For the visual place recognition tasks, we evaluate retrieval performance across different compression levels for each approach. 
Retrieval performance is measured using mean Average Precision at $\mathbf{k}$ (mAP@$\mathbf{k}$), a standard metric for retrieval evaluation. 
Additional results based on the Recall@$\mathbf{k}$ metric are provided in Appendix~\ref{sec:app_recall}. 
For object-centric Monte Carlo localization, we evaluate localization performance using the absolute position error (APE) metric across different compression levels. 
For compression, we measure the memory per element (i.e., per image) transferred over the bandwidth in its compressed form, reported in kilobytes (KB). 
Our approach, \methodname, transmits an LLMZipped text description along with a complementary image feature vector. 
As discussed in Section~\ref{sec:method_ssr}, the size of this feature vector is adaptable, allowing selection of any number of dimensions based on bandwidth or memory constraints. 
The total memory per element in our approach is the sum of the complementary vector's size (depending on the selected number of dimensions) and the size of the LLMZipped text.

\noindent \textbf{Baselines}: We compare three main categories of baselines. The first two categories are specifically tailored for VPR, while the third category is more general and can be applied to both VPR and object-centric Monte Carlo localization.
The first category includes classical compression techniques, such as \textbf{JPEG}~\cite{wallace1991jpeg} and \textbf{JPEG2000}~\cite{skodras2001jpeg2000}. The second category includes neural network-based compression methods focused on image reconstruction, such as Variational Image Compression (\textbf{VIC})~\cite{vicvariationalimagecompressionscale} and Gaussian Mixture Likelihoods (\textbf{GML})~\cite{cheng2020}. 
In both these categories, we assume that the query image is compressed locally, transmitted to the server, decompressed, and subsequently used for feature extraction. For JPEG and JPEG2000, the memory per element is simply the size of the compressed image (in KB) in the respective format. For neural network-based methods, the memory per element corresponds to the size of the compressed latent representation (in KB), which is decompressed on the server side for feature extraction.
\begin{figure*}[ht]
    \centering
    \includegraphics[width=\textwidth]{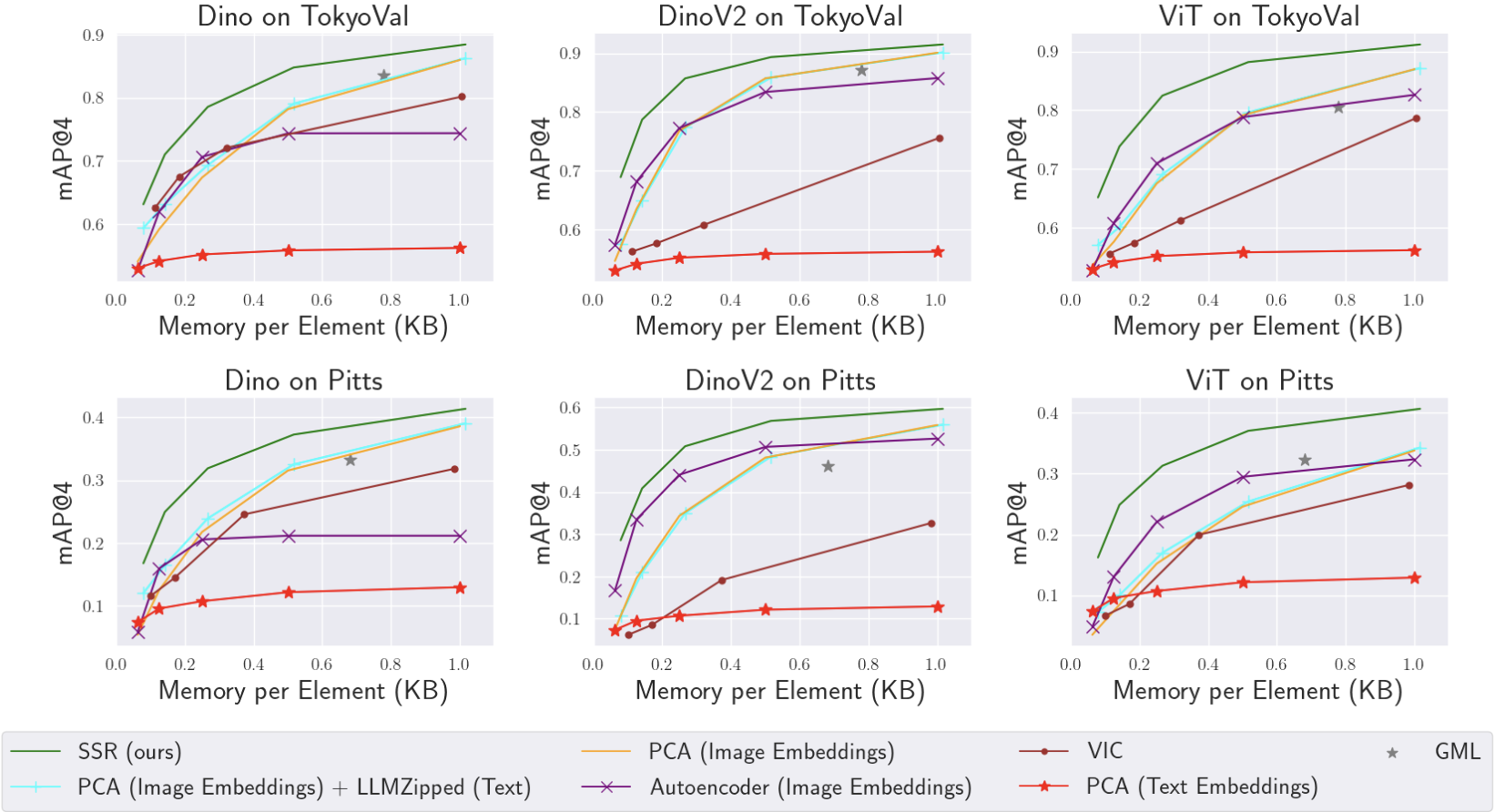} 
    \caption{\textbf{\methodname\ is highly effective for map compression in VPR settings} We show the place recognition performance of all approaches across various compression levels on the TokyoVal and Pittsburgh datasets, using Dino, DinoV2, and ViT embeddings. \methodname\ (green) consistently outperforms all baselines, particularly at smaller memory footprints.}
    \label{fig:exp_main} 
\end{figure*}
The third category assumes feature extraction is performed locally, followed by compression of the extracted features using dimensionality reduction techniques. We evaluate the following baselines in this category:
\begin{enumerate}
    \item \textbf{PCA (Image Embedding)}: Image feature vectors are generated for localization and then compressed using PCA. The memory per element is the size of the PCA-compressed embedding.
    \item \textbf{Autoencoder (Image Embedding)}: Image feature vectors are generated for localization and then compressed using an Autoencoder. The memory per element is the size of the Autoencoder-compressed embedding.
    \item \textbf{PCA (Text Embedding)}: An image caption is generated, its embedding is obtained, which is then compressed using PCA. The memory per element is the size of the PCA-compressed text embedding.
    \item \textbf{PCA (Image Embedding) + LLMZipped (Text)}: Image feature vectors are generated and compressed using PCA. In addition, image captions are generated and compressed using LLMZip. The total memory per element is the sum of the PCA-compressed image vector size and the LLMZipped text size. On the server side, the text is decompressed, its feature vector is generated, concatenated with the PCA-compressed image vector, and used for localization.
\end{enumerate}

\noindent We further evaluate other feature vector compression techniques like quantization in this category. 
Our approach, however, transmits LLMZipped text data with a negligible memory footprint combined with a small complementary image vector. This setup renders all feature vector compression techniques (like quantization, etc.) orthogonal to our method, as the same compression techniques can be applied to the complementary small image vector while the LLMZipped text data has a near-negligible memory footprint. We discuss this in Appendix \ref{app:quantization}

\subsection{Results}\label{sec:exp_results}

\noindent In Fig.~\ref{fig:exp_main}, we compare \methodname\ against the baselines discussed in Sec.~\ref{sec:exp_setup} for the \textbf{VPR} task. 
Note that the performance of JPEG and JPEG2000 was poor; therefore, we omit these results from the main text but include them in Appendix~\ref{sec:app_jpeg}. 
The rows in Fig.~\ref{fig:exp_main} correspond to different datasets, while the columns represent various feature extractors, as described in Sec.~\ref{sec:exp_setup}. 
Our goal is to demonstrate that our approach is compatible with any dataset and feature extractor used for localization. 
\methodname\ (green) consistently outperforms all baselines across all scenarios, achieving on average $2\times$ better compression than competing baselines. 
For example, on the Pittsburgh30K dataset with ViT embeddings, \methodname\ achieves a performance of 0.34 mAP with a memory footprint of only 0.4~KB, whereas the closest baseline, Autoencoders (magenta), requires approximately 1~KB per element. 
Interestingly, we trained different Autoencoder models for each compression level (dimension size). 
In contrast, for \methodname\, we only had to train a single model and then select any desired dimension size at inference, as discussed in Sec.~\ref{sec:method_ssr}.
The results also show that the second category of baselines (VIC (brown) and GML (gray)) perform poorly, as they focus on image reconstruction rather than similarity search. 
Moreover, GML results are shown only for specific compression levels, as it is not easily adaptable to varying compression ratios. 
Finally, \methodname\ surpasses both PCA and Autoencoders in the third category of baselines, highlighting its optimization specifically for localization tasks.
Another notable baseline is ``PCA (Image Embedding) + LLMZipped (Text)" (cyan), which is structurally similar to our method: transmitting a compressed text description along with a complementary image feature vector. 
However, \methodname\ significantly outperforms this baseline, emphasizing the value of \textit{learning} complementary image information rather than relying on independent compression pipelines. Since VPR is an image retrieval problem, our approach also applies to general image retrieval datasets, as discussed in Appendix \ref{app:image_retrieval}.

\noindent In Fig.~\ref{fig:exp_monte_all}, we evaluate \methodname\ on object-centric Monte Carlo localization tasks. 
We present results on two indoor scenes from the Replica dataset (Room0 and Room1) and two outdoor sequences from the KITTI dataset (Seq00 and Seq01). 
For this task, we focus exclusively on the third category of baselines — those based on feature compression rather than full image compression. 
This is because object-centric mapping extracts object crops using approaches such as SAM~\cite{sam}, obtains their features, and fuses features from multiple viewpoints. Baselines like VIC and GML cannot be extended to this setting. Note that we use DINO for object feature extraction to remain consistent with~\cite{sparseloc, conceptgraphs}, although any feature extractor can be used.
The performance is measured in terms of absolute position error (APE) across varying compression levels. 
Consistent with the VPR results, \methodname, outperforms traditional PCA- and Autoencoder-based approaches, demonstrating that our method generalizes effectively to downstream localization tasks beyond retrieval.

\section{Conclusion, Limitations, and Future Work}

In this paper, we proposed \methodname \ that uses text as an alternative modality that captures information complementary to the image data.
It achieves $2\times$ better compression than competing baselines. 
Although \methodname\ saves memory and bandwidth, it is computationally intensive.
During inference, it runs a VLM to generate captions and then LLMZip and a text encoder to compress the text to embeddings, as shown in Fig.\ \ref{fig:pipeline}(B). Thus, \methodname\ offers a trade-off between computation, memory, and bandwidth.
Also, since \methodname\ is based on VLMs and the compressibility of text, it cannot extend to other modalities lacking VLMs, such as inertial measurement units.
Future works include extending our text-based compression technique to support additional vision tasks beyond localization, such as joint image and text compression and reconstruction.
Another area is to optimize LLM prompts \cite{khattab2022demonstrate} to generate captions that fully capture the visual information in images, thus we can completely discard image embeddings in our compression framework.

\begin{figure}
  \centering
  \includegraphics[width=\linewidth]{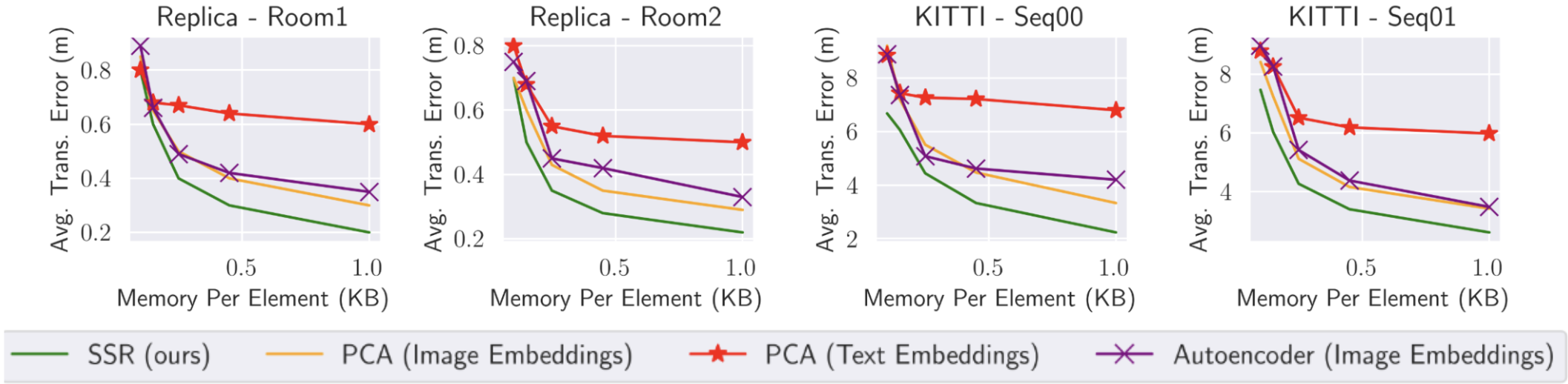} 
  \caption{\textbf{\methodname\ generalizes effectively to Monte Carlo localization tasks.} We show the localization performance (absolute position error) of all approaches across various compression levels on two Replica rooms and two KITTI sequences. \methodname\ (green) consistently achieves lower localization error than competing baselines.}
  \label{fig:exp_monte_all}
\end{figure}


\clearpage


\bibliography{bibtex/external, bibtex/iclr}  

@String(CVPR= {IEEE Conf. Comput. Vis. Pattern Recog.})

@String(ICCV= {Int. Conf. Comput. Vis.})

@String(ECCV= {Eur. Conf. Comput. Vis.})

@String(ICLR = {Int. Conf. Learn. Represent.})

@String(IJCAI = {IJCAI})

@String(CVPR  = {CVPR})

@String(ICCV  = {ICCV})

@String(ECCV  = {ECCV})

@String(ICLR  = {ICLR})

@article{conceptgraphs,
  author    = {Gu, Qiao and Kuwajerwala, Alihusein and Morin, Sacha and Jatavallabhula, {Krishna Murthy} and  Sen, Bipasha and Agarwal, Aditya and Rivera, Corban and Paul, William and Ellis, Kirsty and Chellappa, Rama and Gan, Chuang and {de Melo}, {Celso Miguel} and Tenenbaum, {Joshua B.} and Torralba, Antonio and Shkurti, Florian and Paull, Liam},
  title     = {ConceptGraphs: Open-Vocabulary 3D Scene Graphs for Perception and Planning},
  journal   = {arXiv},
  year      = {2023},
}

@article{sam,
  title={Segment anything},
  author={Kirillov, Alexander and Mintun, Eric and Ravi, Nikhila and Mao, Hanzi and Rolland, Chloe and Gustafson, Laura and Xiao, Tete and Whitehead, Spencer and Berg, Alexander C and Lo, Wan-Yen and others},
  journal={arXiv preprint arXiv:2304.02643},
  year={2023}
}

@article{kusupati2022matryoshka,
  title={Matryoshka representation learning},
  author={Kusupati, Aditya and Bhatt, Gantavya and Rege, Aniket and Wallingford, Matthew and Sinha, Aditya and Ramanujan, Vivek and Howard-Snyder, William and Chen, Kaifeng and Kakade, Sham and Jain, Prateek and others},
  journal={Advances in Neural Information Processing Systems},
  volume={35},
  pages={30233--30249},
  year={2022}
}

@article{valmeekam2023llmzip,
  title={Llmzip: Lossless text compression using large language models},
  author={Valmeekam, Chandra Shekhara Kaushik and Narayanan, Krishna and Kalathil, Dileep and Chamberland, Jean-Francois and Shakkottai, Srinivas},
  journal={arXiv preprint arXiv:2306.04050},
  year={2023}
}

@article{wallace1991jpeg,
  author = {Gregory K. Wallace},
  title = {The JPEG Still Picture Compression Standard},
  journal = {Communications of the ACM},
  volume = {34},
  number = {4},
  pages = {30--44},
  year = {1991},
  publisher = {ACM}
}

@article{omama2024exploiting,
    title={Exploiting Distribution Constraints for Scalable and Efficient Image Retrieval},
    author={Omama, Mohammad and Li, Po-han and Chinchali, Sandeep P},
    journal={arXiv preprint arXiv:2410.07022},
    year={2024}
}

@article{li2024csa,
  title={CSA: Data-efficient Mapping of Unimodal Features to Multimodal Features},
  author={Li, Po-han and Chinchali, Sandeep P and Topcu, Ufuk},
  journal={arXiv preprint arXiv:2410.07610},
  year={2024}
}

@article{theis2017lossy,
  title={Lossy image compression with compressive autoencoders},
  author={Theis, Lucas and Shi, Wenzhe and Cunningham, Andrew and Husz{\'a}r, Ferenc},
  journal={arXiv preprint arXiv:1703.00395},
  year={2017}
}

@article{mishra2022deep,
  title={Deep architectures for image compression: a critical review},
  author={Mishra, Dipti and Singh, Satish Kumar and Singh, Rajat Kumar},
  journal={Signal Processing},
  volume={191},
  pages={108346},
  year={2022},
  publisher={Elsevier}
}

@article{ma2019image,
  title={Image and video compression with neural networks: A review},
  author={Ma, Siwei and Zhang, Xinfeng and Jia, Chuanmin and Zhao, Zhenghui and Wang, Shiqi and Wang, Shanshe},
  journal={IEEE Transactions on Circuits and Systems for Video Technology},
  volume={30},
  number={6},
  pages={1683--1698},
  year={2019},
  publisher={IEEE}
}

@inproceedings{agustsson2019generative,
  title={Generative Adversarial Networks for Extreme Learned Image Compression},
  author={Agustsson, Eirikur and Tschannen, Michael and Mentzer, Fabian and Timofte, Radu and Van Gool, Luc},
  booktitle={Proceedings of the IEEE International Conference on Computer Vision Workshops (ICCVW)},
  pages={221--231},
  year={2019}
}

@inproceedings{toderici2016full,
  title={Full Resolution Image Compression with Recurrent Neural Networks},
  author={Toderici, George and Vincent, Damien and Johnston, Nick and Hwang, Sung Jin and Balle, Johannes and Minnen, David and Shor, Steven and Covell, Michele},
  booktitle={Proceedings of the IEEE Conference on Computer Vision and Pattern Recognition (CVPR)},
  pages={5306--5314},
  year={2016}
}

@article{abdi2010principal,
  title={Principal component analysis},
  author={Abdi, Herv{\'e} and Williams, Lynne J},
  journal={Wiley interdisciplinary reviews: computational statistics},
  volume={2},
  number={4},
  pages={433--459},
  year={2010},
  publisher={Wiley Online Library}
}

@book{gersho2012vector,
  title={Vector quantization and signal compression},
  author={Gersho, Allen and Gray, Robert M},
  volume={159},
  year={2012},
  publisher={Springer Science \& Business Media}
}

@inproceedings{chen2024internvl,
  title={Internvl: Scaling up vision foundation models and aligning for generic visual-linguistic tasks},
  author={Chen, Zhe and Wu, Jiannan and Wang, Wenhai and Su, Weijie and Chen, Guo and Xing, Sen and Zhong, Muyan and Zhang, Qinglong and Zhu, Xizhou and Lu, Lewei and others},
  booktitle={Proceedings of the IEEE/CVF Conference on Computer Vision and Pattern Recognition},
  pages={24185--24198},
  year={2024}
}

@misc{dai2023instructblipgeneralpurposevisionlanguagemodels,
      title={InstructBLIP: Towards General-purpose Vision-Language Models with Instruction Tuning}, 
      author={Wenliang Dai and Junnan Li and Dongxu Li and Anthony Meng Huat Tiong and Junqi Zhao and Weisheng Wang and Boyang Li and Pascale Fung and Steven Hoi},
      year={2023},
      eprint={2305.06500},
      archivePrefix={arXiv},
      primaryClass={cs.CV},
      url={https://arxiv.org/abs/2305.06500}, 
}

@article{skodras2001jpeg2000,
  title={JPEG 2000 image coding system: digital compression for still images},
  author={Skodras, Athanassios and Christopoulos, Charilaos and Ebrahimi, Touradj},
  journal={IEEE Transactions on Consumer Electronics},
  volume={46},
  number={4},
  pages={919-929},
  year={2001},
  publisher={IEEE}
}

@article{khattab2022demonstrate,
  title={Demonstrate-Search-Predict: Composing Retrieval and Language Models for Knowledge-Intensive {NLP}},
  author={Khattab, Omar and Santhanam, Keshav and Li, Xiang Lisa and Hall, David and Liang, Percy and Potts, Christopher and Zaharia, Matei},
  journal={arXiv preprint arXiv:2212.14024},
  year={2022}
}

@misc{cheng2020,
      title={Learned Image Compression with Discretized Gaussian Mixture Likelihoods and Attention Modules}, 
      author={Zhengxue Cheng and Heming Sun and Masaru Takeuchi and Jiro Katto},
      year={2020},
      eprint={2001.01568},
      archivePrefix={arXiv},
      primaryClass={eess.IV},
      url={https://arxiv.org/abs/2001.01568}, 
}

@misc{vicvariationalimagecompressionscale,
      title={Variational image compression with a scale hyperprior}, 
      author={Johannes Ballé and David Minnen and Saurabh Singh and Sung Jin Hwang and Nick Johnston},
      year={2018},
      eprint={1802.01436},
      archivePrefix={arXiv},
      primaryClass={eess.IV},
      url={https://arxiv.org/abs/1802.01436}, 
}

@article{touvron2023llama,
  title={LLaMA: Open and Efficient Foundation Language Models},
  author={Touvron, Hugo and Lavril, Thibaut and Izacard, Gautier and Martinet, Xavier and Lachaux, Marie-Anne and Lacroix, Timothée and Rozière, Baptiste and Goyal, Naman and Hambro, Eric and Azhar, Moin and Rodriguez, Aurelien and Joulin, Armand and Grave, Edouard and Lample, Guillaume},
  journal={arXiv preprint arXiv:2302.13971},
  year={2023}
}

@article{composedret1,
author = {Ge, Hongfei and Jiang, Yuanchun and Sun, Jianshan and Yuan, Kun and Liu, Yezheng},
title = {LLM-enhanced Composed Image Retrieval: An Intent Uncertainty-aware Linguistic-Visual Dual Channel Matching Model},
year = {2024},
publisher = {Association for Computing Machinery},
address = {New York, NY, USA},
issn = {1046-8188},
url = {https://doi.org/10.1145/3699715},
doi = {10.1145/3699715},
note = {Just Accepted},
journal = {ACM Trans. Inf. Syst.},
month = oct
}

@inproceedings{composedret2,
author = {Yang, Zhenyu and Xue, Dizhan and Qian, Shengsheng and Dong, Weiming and Xu, Changsheng},
title = {LDRE: LLM-based Divergent Reasoning and Ensemble for Zero-Shot Composed Image Retrieval},
year = {2024},
isbn = {9798400704314},
publisher = {Association for Computing Machinery},
address = {New York, NY, USA},
url = {https://doi.org/10.1145/3626772.3657740},
doi = {10.1145/3626772.3657740},
booktitle = {Proceedings of the 47th International ACM SIGIR Conference on Research and Development in Information Retrieval},
pages = {80–90},
numpages = {11},
location = {Washington DC, USA},
series = {SIGIR '24}
}

@inproceedings{kitti,
  title={Are we ready for Autonomous Driving? The KITTI Vision Benchmark Suite},
  author={Geiger, Andreas and Lenz, Philip and Urtasun, Raquel},
  booktitle={Conference on Computer Vision and Pattern Recognition (CVPR)},
  year={2012}
}

@article{replica,
  title={The Replica dataset: A digital replica of indoor spaces},
  author={Straub, Julian and Whelan, Thomas and Ma, Lingni and Chen, Yufeng and Wijmans, Erik and Green, Spencer and Engel, Jakob and Mur-Artal, Raul and Ren, Changhan and Verma, Rohit and others},
  journal={arXiv preprint arXiv:1906.05797},
  year={2019}
}

@article{sparseloc,
  title={SparseLoc: Sparse Open-Set Landmark-based Global Localization for Autonomous Navigation},
  author={Paul, Pranjal and Bhat, Vineeth and Salian, Tejas and Omama, Mohammad and Jatavallabhula, Krishna Murthy and Arulselvan, Naveen and Krishna, K Madhava},
  journal={arXiv preprint arXiv:2503.23465},
  year={2025}
}

@inproceedings{liu2023llava,
    author = {Liu, Haotian and Li, Chunyuan and Wu, Qingyang and Lee, Yong Jae},
    booktitle = {Advances in Neural Information Processing Systems},
    editor = {A. Oh and T. Naumann and A. Globerson and K. Saenko and M. Hardt and S. Levine},
    pages = {34892--34916},
    publisher = {Curran Associates, Inc.},
    title = {Visual Instruction Tuning},
    url = {https://proceedings.neurips.cc/paper_files/paper/2023/file/6dcf277ea32ce3288914faf369fe6de0-Paper-Conference.pdf},
    volume = {36},
    year = {2023}
}

@misc{li2023blip2,
      title={BLIP-2: Bootstrapping Language-Image Pre-training with Frozen Image Encoders and Large Language Models}, 
      author={Junnan Li and Dongxu Li and Silvio Savarese and Steven Hoi},
      year={2023},
      eprint={2301.12597},
      archivePrefix={arXiv},
      primaryClass={cs.CV}
}

@inproceedings{radford2021learning,
  title={Learning transferable visual models from natural language supervision},
  author={Radford, Alec and Kim, Jong Wook and Hallacy, Chris and Ramesh, Aditya and Goh, Gabriel and Agarwal, Sandhini and Sastry, Girish and Askell, Amanda and Mishkin, Pamela and Clark, Jack and others},
  booktitle={International conference on machine learning},
  pages={8748--8763},
  year={2021},
  organization={PMLR}
}

@inproceedings{pittsdataset,
  title={Visual place recognition with repetitive structures},
  author={Torii, Akihiko and Sivic, Josef and Pajdla, Tomas and Okutomi, Masatoshi},
  booktitle={Proceedings of the IEEE conference on computer vision and pattern recognition},
  pages={883--890},
  year={2013}
}

@inproceedings{tokyodataset,
  title={24/7 place recognition by view synthesis},
  author={Torii, Akihiko and Arandjelovic, Relja and Sivic, Josef and Okutomi, Masatoshi and Pajdla, Tomas},
  booktitle={Proceedings of the IEEE conference on computer vision and pattern recognition},
  pages={1808--1817},
  year={2015}
}

@misc{dino,
      title={Emerging Properties in Self-Supervised Vision Transformers}, 
      author={Mathilde Caron and Hugo Touvron and Ishan Misra and Hervé Jégou and Julien Mairal and Piotr Bojanowski and Armand Joulin},
      year={2021},
      eprint={2104.14294},
      archivePrefix={arXiv},
      primaryClass={cs.CV}
}

@misc{dinov2,
      title={DINOv2: Learning Robust Visual Features without Supervision}, 
      author={Maxime Oquab and Timothée Darcet and Théo Moutakanni and Huy Vo and Marc Szafraniec and Vasil Khalidov and Pierre Fernandez and Daniel Haziza and Francisco Massa and Alaaeldin El-Nouby and Mahmoud Assran and Nicolas Ballas and Wojciech Galuba and Russell Howes and Po-Yao Huang and Shang-Wen Li and Ishan Misra and Michael Rabbat and Vasu Sharma and Gabriel Synnaeve and Hu Xu and Hervé Jegou and Julien Mairal and Patrick Labatut and Armand Joulin and Piotr Bojanowski},
      year={2024},
      eprint={2304.07193},
      archivePrefix={arXiv},
      primaryClass={cs.CV}
}

@misc{vit,
      title={An Image is Worth 16x16 Words: Transformers for Image Recognition at Scale}, 
      author={Alexey Dosovitskiy and Lucas Beyer and Alexander Kolesnikov and Dirk Weissenborn and Xiaohua Zhai and Thomas Unterthiner and Mostafa Dehghani and Matthias Minderer and Georg Heigold and Sylvain Gelly and Jakob Uszkoreit and Neil Houlsby},
      year={2021},
      eprint={2010.11929},
      archivePrefix={arXiv},
      primaryClass={cs.CV}
}

@misc{dinov2data:lvd142m,
  title = {LVD-142M: A Curated Dataset for Self-Supervised Learning},
  author = {Oquab, Maxime and Darcet, Timothée and Moutakanni, Theo and Vo, Huy V. and Szafraniec, Marc and Khalidov, Vasil and Fernandez, Pierre and Haziza, Daniel and Massa, Francisco and El-Nouby, Alaaeldin and Howes, Russell and Huang, Po-Yao and Xu, Hu and Sharma, Vasu and Li, Shang-Wen and Galuba, Wojciech and Rabbat, Mike and Assran, Mido and Ballas, Nicolas and Synnaeve, Gabriel and Misra, Ishan and Jegou, Herve and Mairal, Julien and Labatut, Patrick and Joulin, Armand and Bojanowski, Piotr},
  year = {2023},
  note = {Utilized in the pretraining of DINOv2 models},
  howpublished = {\url{https://github.com/facebookresearch/dinov2}},
}

@article{russakovsky2015imagenet,
  title={Imagenet large scale visual recognition challenge},
  author={Russakovsky, Olga and Deng, Jia and Su, Hao and Krause, Jonathan and Satheesh, Sanjeev and Ma, Sean and Huang, Zhiheng and Karpathy, Andrej and Khosla, Aditya and Bernstein, Michael and others},
  journal={International journal of computer vision},
  volume={115},
  pages={211--252},
  year={2015},
  publisher={Springer}
}

@article{kingma2013auto,
  title={Auto-Encoding Variational Bayes},
  author={Kingma, Diederik P and Welling, Max},
  journal={arXiv preprint arXiv:1312.6114},
  year={2013}
}

@inproceedings{cvpr-context-sim-distill,
  title={Contextual similarity distillation for asymmetric image retrieval},
  author={Wu, Hui and Wang, Min and Zhou, Wengang and Li, Houqiang and Tian, Qi},
  booktitle={Proceedings of the IEEE/CVF Conference on Computer Vision and Pattern Recognition},
  pages={9489--9498},
  year={2022}
}

@article{li2024task,
  title={Task-aware Distributed Source Coding under Dynamic Bandwidth},
  author={Li, Po-han and Ankireddy, Sravan Kumar and Zhao, Ruihan Philip and Nourkhiz Mahjoub, Hossein and Moradi Pari, Ehsan and Topcu, Ufuk and Chinchali, Sandeep and Kim, Hyeji},
  journal={Advances in Neural Information Processing Systems},
  volume={36},
  year={2023}
}

@inproceedings{cosplace,
  title={Rethinking visual geo-localization for large-scale applications},
  author={Berton, Gabriele and Masone, Carlo and Caputo, Barbara},
  booktitle={Proceedings of the IEEE/CVF Conference on Computer Vision and Pattern Recognition},
  pages={4878--4888},
  year={2022}
}

@inproceedings{dml1,
  title={Deep metric learning via lifted structured feature embedding},
  author={Oh Song, Hyun and Xiang, Yu and Jegelka, Stefanie and Savarese, Silvio},
  booktitle={Proceedings of the IEEE conference on computer vision and pattern recognition},
  pages={4004--4012},
  year={2016}
}

@inproceedings{dml2facenet,
  title={Facenet: A unified embedding for face recognition and clustering},
  author={Schroff, Florian and Kalenichenko, Dmitry and Philbin, James},
  booktitle={Proceedings of the IEEE conference on computer vision and pattern recognition},
  pages={815--823},
  year={2015}
}

@article{dml3,
  title={Learning deep embeddings with histogram loss},
  author={Ustinova, Evgeniya and Lempitsky, Victor},
  journal={Advances in neural information processing systems},
  volume={29},
  year={2016}
}

@article{dml4-npair,
  title={Improved deep metric learning with multi-class n-pair loss objective},
  author={Sohn, Kihyuk},
  journal={Advances in neural information processing systems},
  volume={29},
  year={2016}
}

@inproceedings{dml5-msl,
  title={Multi-similarity loss with general pair weighting for deep metric learning},
  author={Wang, Xun and Han, Xintong and Huang, Weilin and Dong, Dengke and Scott, Matthew R},
  booktitle={Proceedings of the IEEE/CVF conference on computer vision and pattern recognition},
  pages={5022--5030},
  year={2019}
}

@article{dml-survey,
  title={Deep Metric Learning: A Survey},
  author={Wang, Jiang and Song, Yang and Leung, Thomas and Rosenberg, Chuck and Wang, Jingbin and Philbin, James and Chen, Bo and Wu, Ying},
  journal={arXiv preprint arXiv:1706.09720},
  year={2017}
}

@inproceedings{dml-contrastive,
  title={Dimensionality reduction by learning an invariant mapping},
  author={Hadsell, Raia and Chopra, Sumit and LeCun, Yann},
  booktitle={2006 IEEE computer society conference on computer vision and pattern recognition (CVPR'06)},
  volume={2},
  pages={1735--1742},
  year={2006},
  organization={IEEE}
}

@inproceedings{vpr-intro,
  title={Where is your place, visual place recognition?},
  author={Garg, Sourav and Fischer, Tobias and Milford, Michael},
  booktitle={IJCAI},
  volume={8},
  pages={4416--4425},
  year={2021}
}

@inproceedings{vpr1,
  title={Self-supervising fine-grained region similarities for large-scale image localization},
  author={Ge, Yixiao and Wang, Haibo and Zhu, Feng and Zhao, Rui and Li, Hongsheng},
  booktitle={Computer Vision--ECCV 2020: 16th European Conference, Glasgow, UK, August 23--28, 2020, Proceedings, Part IV 16},
  pages={369--386},
  year={2020},
  organization={Springer}
}

@article{vpr2,
  title={Visual geo-localization with self-supervised representation learning},
  author={Xiao, Jiuhong and Zhu, Gao and Loianno, Giuseppe},
  journal={arXiv preprint arXiv:2308.00090},
  year={2023}
}

@inproceedings{vpr3,
  title={Data-efficient large scale place recognition with graded similarity supervision},
  author={Leyva-Vallina, Mar{\'\i}a and Strisciuglio, Nicola and Petkov, Nicolai},
  booktitle={Proceedings of the IEEE/CVF Conference on Computer Vision and Pattern Recognition},
  pages={23487--23496},
  year={2023}
}

@article{romero2014fitnets,
  title={FitNets: Hints for thin deep nets},
  author={Romero, Adriana and Ballas, Nicolas and Kahou, Samira Ebrahimi and Chassang, Antoine and Gatta, Carlo and Bengio, Yoshua},
  journal={arXiv preprint arXiv:1412.6550},
  year={2014}
}

@inproceedings{dist4,
  title={Paying more attention to attention: Improving the performance of convolutional neural networks via attention transfer},
  author={Zagoruyko, Sergey and Komodakis, Nikos},
  booktitle={International Conference on Learning Representations (ICLR)},
  year={2017}
}

@inproceedings{tung2019similarity,
  title={Similarity-preserving knowledge distillation},
  author={Tung, Frederick and Mori, Greg},
  booktitle={Proceedings of the IEEE International Conference on Computer Vision (ICCV)},
  year={2019}
}

@inproceedings{tian2019contrastive,
  title={Contrastive representation distillation},
  author={Tian, Yonglong and Krishnan, Dilip and Isola, Phillip},
  booktitle={International Conference on Learning Representations (ICLR)},
  year={2020}
}

@inproceedings{park2019relational,
  title={Relational knowledge distillation},
  author={Park, Wonpyo and Kim, Dongju and Lu, Yan and Cho, Minsu},
  booktitle={Proceedings of the IEEE Conference on Computer Vision and Pattern Recognition (CVPR)},
  year={2019}
}

@article{dist1,
  title={Distilling the knowledge in a neural network},
  author={Hinton, Geoffrey and Vinyals, Oriol and Dean, Jeff},
  journal={arXiv preprint arXiv:1503.02531},
  year={2015}
}

@inproceedings{dist2,
  title={Model compression},
  author={Buciluǎ, Cristian and Caruana, Rich and Niculescu-Mizil, Alexandru},
  booktitle={Proceedings of the 12th ACM SIGKDD international conference on Knowledge discovery and data mining},
  pages={535--541},
  year={2006}
}

@inproceedings{dist3,
  title={Do deep nets really need to be deep?},
  author={Ba, Jimmy and Caruana, Rich},
  booktitle={Advances in neural information processing systems},
  pages={2654--2662},
  year={2014}
}

@inproceedings{dist5,
  title={Your classifier is secretly an energy based model and you should treat it like one},
  author={Zhang, Chiyuan and Bengio, Samy and Hardt, Moritz and Recht, Benjamin and Vinyals, Oriol},
  booktitle={International Conference on Learning Representations (ICLR)},
  year={2019}
}

@inproceedings{dist6,
  title={Learning deep representations with probabilistic knowledge transfer},
  author={Passalis, Nikolaos and Tefas, Anastasios},
  booktitle={Proceedings of the European Conference on Computer Vision (ECCV)},
  pages={268--284},
  year={2018}
}

\appendix

\section{Appendix}

\subsection{Extending \methodname\ to General Image Retrieval Datasets} \label{app:image_retrieval}
\begin{figure*}[h]
    \centering
    \includegraphics[width=\textwidth]{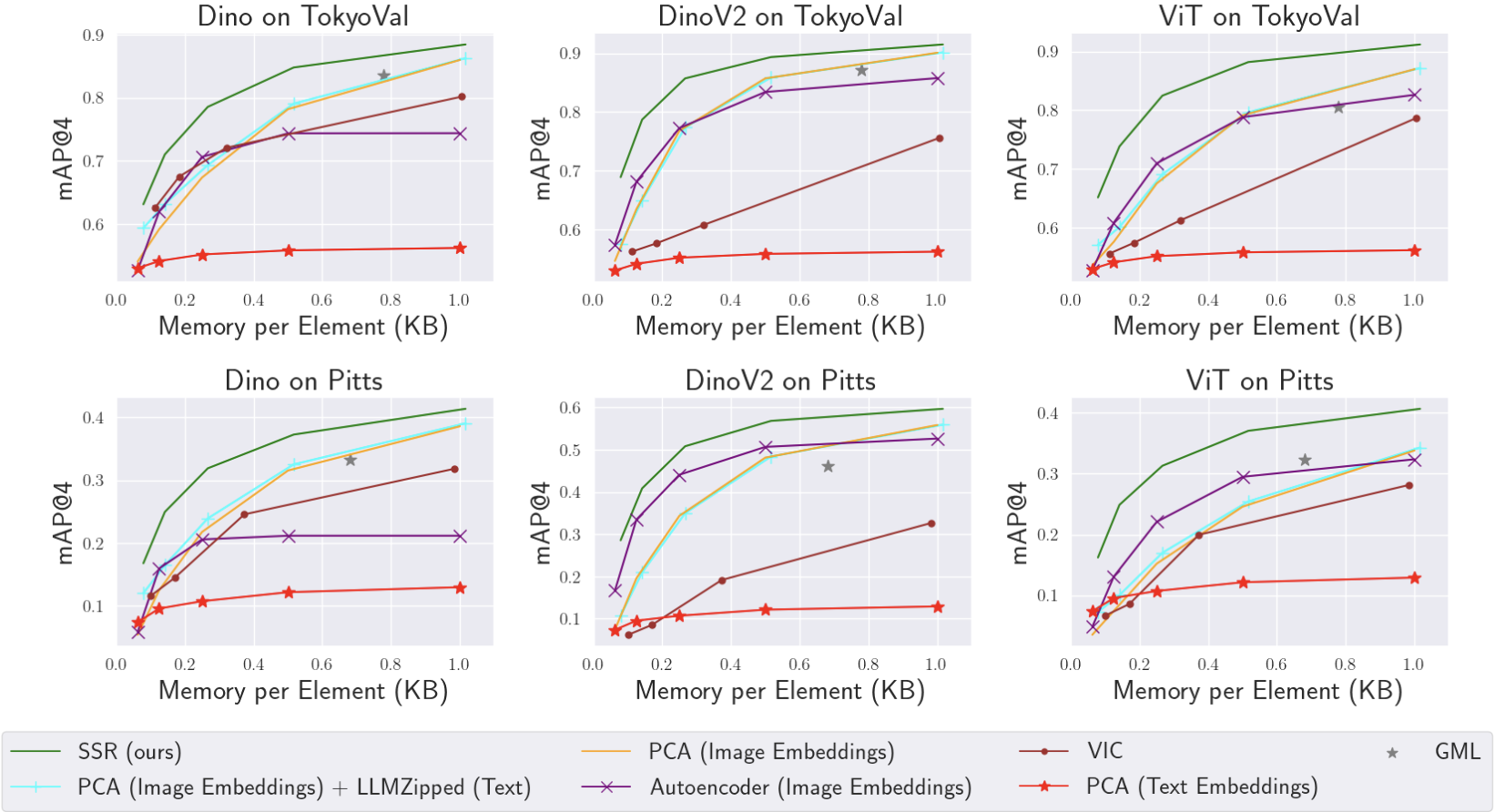} 
    \caption{\textbf{\methodname\ is highly effective for compression in general image retrieval settings as well.} We show the image retrieval performance of all approaches across various compression levels on the InShop and SoP datasets, using Dino, DinoV2, and ViT embeddings. \methodname\ (green) consistently outperforms all baselines, particularly at smaller memory footprints.}
    \label{fig:exp_main_ir} 
\end{figure*}

Since VPR is an image retrieval problem, we can easily show that our proposed solution works well with general image retrieval datasets as well as discussed in Fig. \ref{fig:exp_main_ir}.

\subsection{\methodfullname\ in a Federated Setting (\methodnamefl) } \label{app:method_ssrfl}

\textbf{Motivation:} Localization applications often involve privacy-sensitive data distributed across multiple systems. For example, a tourist in a new city may wish to identify landmarks via reverse image search; however, the reference data may be stored across multiple servers managed by different entities, each unwilling to share their proprietary data for training a joint compression function. This scenario introduces both privacy and bandwidth constraints.
To address these challenges, we extend \methodfullname\ (\methodname) to a Federated Learning (FL) setup, \methodnamefl, enabling privacy-preserving learning of complementary embeddings. \methodnamefl\ meets the requirements of distributed map compression applications by allowing nodes to independently learn and share model updates based on their proprietary data.

\begin{figure*}[ht]
    \centering
    \includegraphics[width=\textwidth]{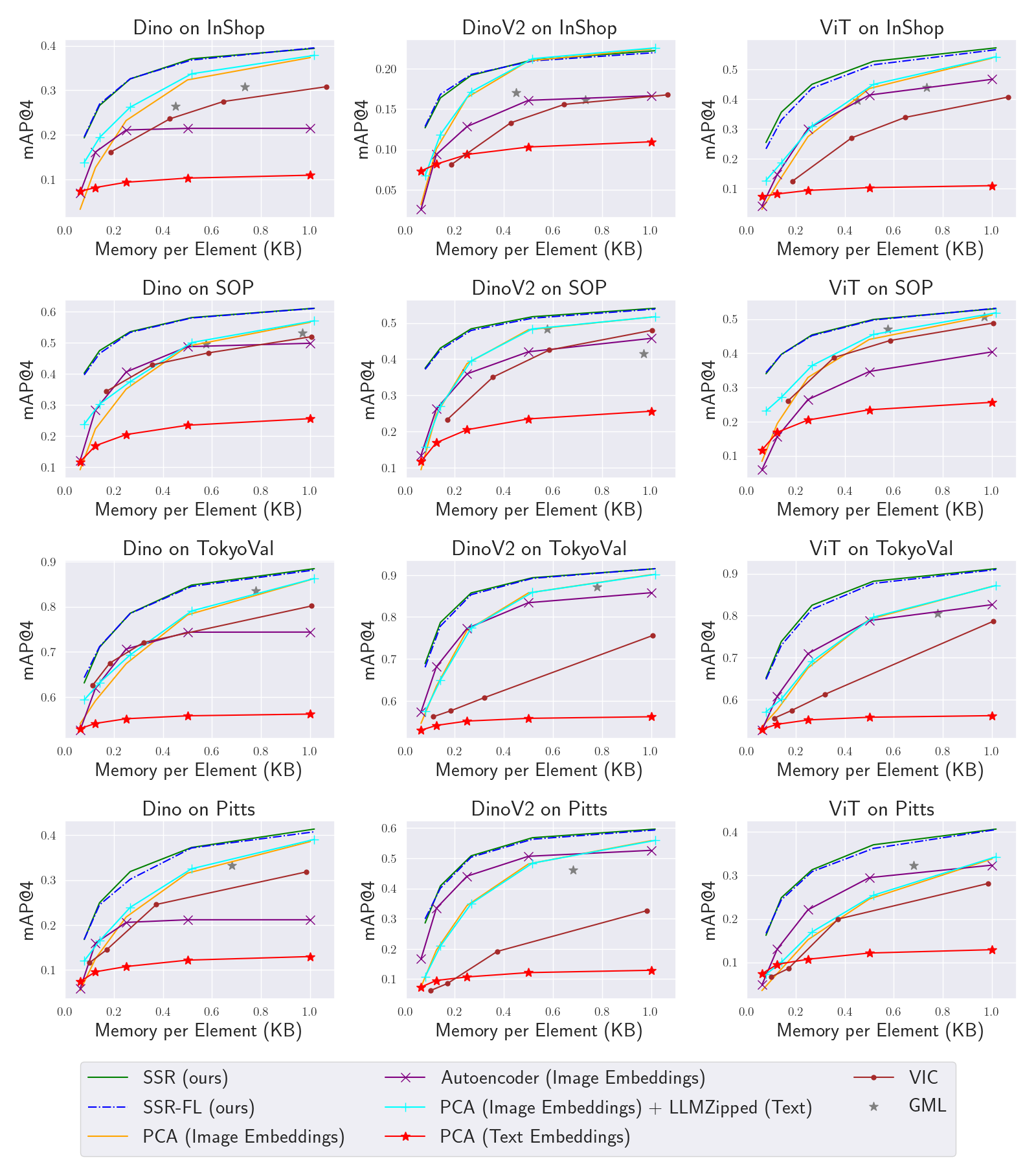} 
    \caption{\textbf{\methodnamefl\ is as effective as \methodname\ for map compression in VPR settings} We show the place recognition performance of all approaches across various compression levels on the TokyoVal and Pittsburgh datasets, using Dino, DinoV2, and ViT embeddings. \methodname\ (green) consistently outperforms all baselines, while \methodnamefl\ closely approximates \methodname.}
    \label{fig:exp_main_fl} 
\end{figure*}

\noindent \textbf{FL Setup:} In our FL setup, we assume there are $A$ nodes or agents, each with access to its own distinct subset of the training data. These nodes collaborate to train a single \methodnamefl\ model by periodically sharing updates to the model parameters, rather than their raw data. Throughout our experiments, we fix $A$ to $4$ without loss of generality.

\noindent \textbf{Training Process:} During each training round, each node $a \in A$ locally optimizes the \methodfullname\ loss to ensure that its local similarity space, constructed with complementary embeddings, closely approximates the full similarity space. Each node $a$ individually computes its similarity matrices and complementary embeddings based on its local data,  with $N_a$ images, using the neural network $\mathcal{G}(\cdot; \phi_a)$, where $\phi_a$ represents the model parameters specific to node $a$. 

For each subset of dimensions $c \in \mathcal{C}$, node $a$ computes the student similarity space $\tilde{\mathcal{S}}^c_a$ by combining the complementary embeddings $\hat{z}^{1:c}_a$ with the local text embeddings. The KL divergence loss $l^c_a$ for each dimension subset $c$ is calculated as:
\begin{equation}
l^c_a = \sum_{j=1}^{N_a} \mathrm{D}_{\mathrm{KL}}(\tilde{\mathcal{S}}^c_{a,j} \| \mathcal{S}_{a,j}),
\end{equation}
where $\mathcal{S}_{a}$ denotes the full similarity space computed for node $a$. The total loss $L_a$ for node $a$ across all embedding dimensions in $\mathcal{C}$ is given by:
\begin{equation}
L_{{\methodname}_a} = \sum_{c \in \mathcal{C}} l^c_a = \sum_{c \in \mathcal{C}} \sum_{j=1}^{N_a} \mathrm{D}_{\mathrm{KL}}(\tilde{\mathcal{S}}^c_{a,j} \| \mathcal{S}_{a,j}).
\end{equation}
Each node $a$ then updates its parameters $\phi_a$ by minimizing $L_{{\methodname}_a}$.
Once these local updates are completed, each node sends its updated parameters $\phi_a$ to the central server, which aggregates the updates by averaging them as:
\begin{equation}
\phi = \frac{1}{A} \sum_{a=1}^{A} \phi_a,
\end{equation}
where $A$ is the total number of nodes. The refined global parameters $\phi$ are then distributed back to each node for the next training round. This iterative process enables our federated \methodnamefl\ model to capture complementary information across different nodes while preserving data privacy. 

In the experiments Fig. \ref{fig:exp_main_fl}, we show that \methodnamefl\ achieves performance close to \methodname, making it well-suited for distributed settings. We argue that this is due to \methodname's training data efficiency, which allows it to be trained in distributed environments efficiently. Since \methodname\ includes an VLM based caption generation, much of the information is already captured in text format, enabling it to focus only on ``complementary" information—an easier task than dimensionality reduction via autoencoders, which must learn reconstruction from the ground up. We further compare the data efficiency of \methodname\ and autoencoders in Appendix \ref{sec:app_dataeff}.

\subsection{Data Efficient Training Comparison } \label{app:dataeff}

In the experiments (Sec.~\ref{sec:exp_results}), we demonstrate that \methodnamefl\ achieves performance comparable to \methodname, making it well-suited for distributed settings. This is attributed to \methodname's sample efficiency in learning ``complementary image features.'' 
To further illustrate this, we compare the data efficiency of \methodname\ with Autoencoders (AE). Using random small subsets of training data, we analyze the performance drop in VPR (mAP@$k$). The results are summarized in Table~\ref{tab:dataeff}. Notably, when using only 25\% of the training data, the performance of AE drops by 12\% (from 0.31 to 0.19), whereas \methodname\ experiences a smaller decline of 6\%. These results are shown using the Pittsburgh dataset with the ViT embedding, maintaining a fixed memory per element for both approaches.

We attribute this data efficiency primarily to the use of the VLM-based caption generation, whereby much of the information is already captured in a text format. Therefore, making the learning process of ``complementary features'' easier.

\begin{table}[ht]
\centering
\begin{tabular}{lcccc}
\toprule
             & AE       & \methodname\  \\ \midrule
25\% Data    &  0.19 &  \textbf{0.31}      \\
50\% Data    &  0.22 &  \textbf{0.33}      \\
75\% Data    &  0.27 &  \textbf{0.35}      \\
100\% Data    &  0.31 &  \textbf{0.37}       \\ \bottomrule
\end{tabular}
\caption{\textbf{Performance ( mAP@$k$)  comparison between \methodname\ and Autoencoders (AE) when using subsets of training data.} AE shows a 12\% performance drop (from 0.31 to 0.19) with 25\% training data, while \methodname\ exhibits a smaller drop of 6\%.}
\label{tab:dataeff}
\end{table}

\subsection{Comparison with Quantization}\label{app:quantization}
\noindent Additionally, we evaluate quantization in this category. 
Our approach, however, transmits LLMZipped text data with a negligible memory footprint combined with a small complementary image vector. This setup renders all feature vector compression techniques (like quantization) orthogonal to our method, as the same compression techniques can be applied to the complementary small image vector, while the LLMZipped text data has near near-negligible memory footprint.
\begin{figure}[t]
    \centering
    \includegraphics[width=0.6\linewidth]{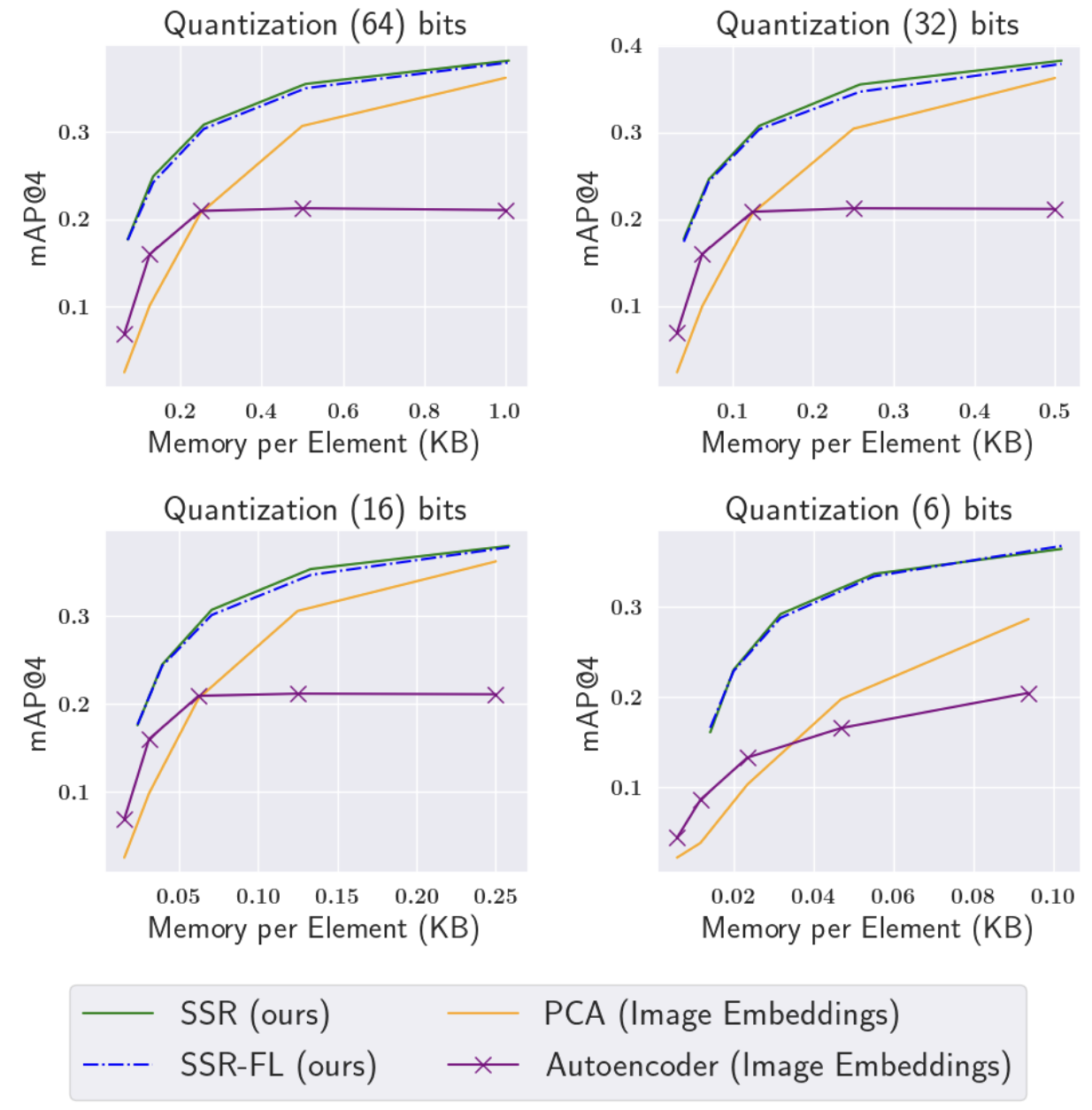} 
    \caption{\textbf{\methodname\ and \methodnamefl\ are orthogonal to quantization}. We compare the performance of various baselines across different quantization levels. Our approach outperforms all baselines at each quantization level. While methods such as PCA and Autoencoders experience significant performance drops at 6-bit quantization, \methodname\ and \methodnamefl\ maintain stable performance.
    }
    \label{fig:exp_quant} 
\end{figure}


\subsection{Additional Results on the Recall Metric} \label{sec:app_recall}
\begin{figure}[ht]
  \centering
  \includegraphics[width=0.7\linewidth]{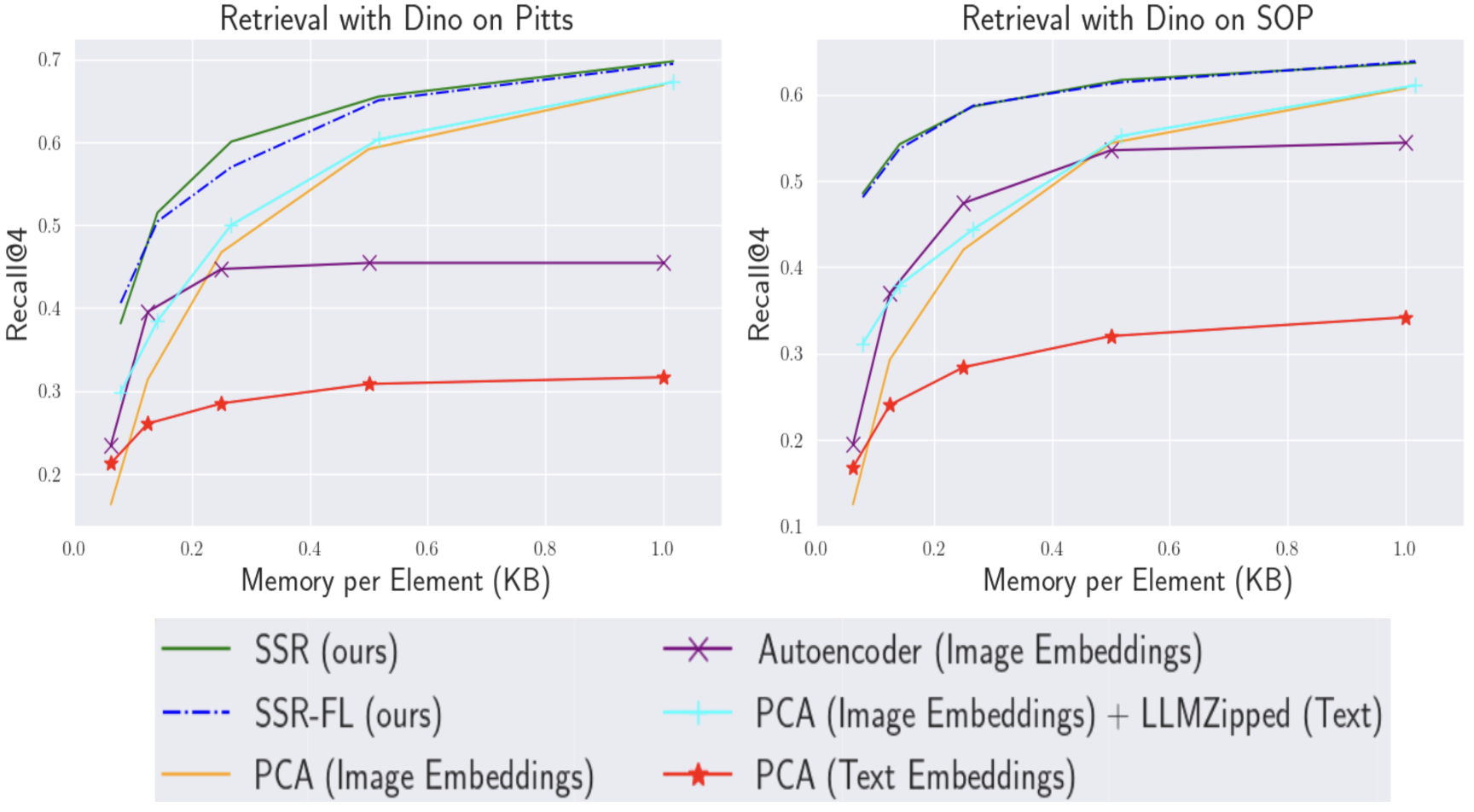}
  \caption{\textbf{Additional results on the Recall@$k$ metric}. We see the same trends as in the main paper and we outperform all other baselines on the Recall@$k$ metric}
  \label{fig:app_recall}
\end{figure}

In the experiments (Sec.~\ref{sec:exp_results}), we showed that our approach outperforms all other baselines on the mAP@$k$. However, other metrics like Recall@$k$ are also common in the VPR literature. In Fig. \ref{fig:app_recall}, we show that the same trends are observed as in Sec. \ref{sec:exp_results} on the Recall@$k$ metric as well. We demonstrate the VPR performance on two datasets using the Dino embedding.

\subsection{Additional Results with JPEG and JPEG2000} \label{sec:app_jpeg}
In the experiments (Sec.~\ref{sec:exp_results}), we did not report the performance of JPEG and JPEG 2000 as they were not comparable to the other baselines. Table~\ref{tab:jpeg_comparison} presents the memory footprint required to achieve a given VPR performance. As shown, \methodname\ requires approximately 10$\times$ less data to achieve the same VPR performance as JPEG and JPEG 2000.

\begin{table}[ht]
\centering
\begin{tabular}{lcccc}
\toprule
             & JPEG       & JPEG 2000   & \methodname\  \\ \midrule
map@4 = 0.2    &  0.1 &  0.09       &  \textbf{0.01}      \\
map@4 = 0.4    &  1.1 &  0.9       &  \textbf{0.1}      \\ \bottomrule
\end{tabular}
\caption{Memory footprint of JPEG, JPEG 2000, and \methodname\ required for given VPR performances.}
\label{tab:jpeg_comparison}
\end{table}

\subsection{Other Details} \label{sec:app_hyperparam}
All baselines requiring training were trained for 5 epochs with a learning rate of $10^{-4}$. For \methodnamefl, we utilized 4 nodes in a federated learning (FL) setup. The training data was randomly split across these 4 nodes. We assumed the presence of a central server to orchestrate the FL training process. However, this central server assumption is not critical to our approach, as \methodnamefl\ can also be trained in a decentralized FL setup.

\end{document}